\documentclass[conference]{IEEEtran}
\usepackage{times}
\usepackage{microtype}
\usepackage[T1]{fontenc}

\usepackage[numbers]{natbib}
\usepackage{multicol}
\usepackage[bookmarks=true]{hyperref}

\usepackage{microtype}
\usepackage{graphicx}
\usepackage{subfigure}
\usepackage{booktabs}
\usepackage{tikz}
\usetikzlibrary{scopes,patterns,arrows.meta,positioning,fit,calc,backgrounds}
\usepackage{xcolor}
\let\labelindent\relax
\usepackage[inline]{enumitem}
\usepackage{amsmath}
\usepackage{amssymb}
\usepackage{mathtools}
\usepackage{amsthm}

\theoremstyle{plain}

\theoremstyle{definition}

\theoremstyle{remark}

\hypersetup{
  pdftitle={KlaskTron: An Open-Source Platform for Physical Adversarial Multi-Agent RL},
  pdfauthor={Aswin Ramachandran, Jona Schulz, Maurus Derungs, Carlo Angelini, Tobias Meier, Raffaello D'Andrea},
  pdfsubject={Robotics},
  pdfkeywords={MARL;Benchmark;Sim-to-Real;Robotics},
}

\begin{document}

\title{KlaskTron: An Open-Source Platform for Physical Adversarial Multi-Agent RL}

\author{\IEEEauthorblockN{Aswin Ramachandran, Jona Schulz, Maurus Derungs,\\ Carlo Angelini, Tobias Meier, Raffaello D'Andrea}
\IEEEauthorblockA{Institute for Dynamic Systems and Control (IDSC), ETH Z\"urich, Switzerland\\
Email: \texttt{aramachandra@ethz.ch}}}

\maketitle
\begin{abstract}
Progress in robot learning, particularly physical multi-agent reinforcement learning (MARL), is currently challenged by a limited availability of accessible, standardized benchmarks. 
While simulation-based MARL has produced remarkable emergent behaviors from coordinated team play to complex tool use, translating these advances to physical systems remains difficult. 
A key barrier is infrastructural: physical platforms are often prohibitively expensive, require specialized facilities, or lack open mechanisms and designs for reproducibility. 
We introduce KlaskTron, an open-source, low-cost (<\$2500 USD), desktop-scale robotic testbed for adversarial MARL based on the dynamic tabletop game KLASK. 
The platform features dual CoreXY gantries with high-torque brushless motors, enabling the high-speed, precise manipulation required for competitive play. 
Crucially, we provide a complete ecosystem: hardware designs (CAD, BOM), a GPU-native digital twin in NVIDIA Isaac Lab for high-fidelity modeling and simulation, and a validated sim-to-real baseline using a learned neural actuator model. 
We demonstrate that this baseline enables zero-shot policy transfer, with trained agents exhibiting emergent adversarial behaviors. To ensure full reproducibility of our results, all project assets are released publicly: hardware CAD/BOM and assembly documentation at \url{https://idscethzurich.github.io/klask_hardware/} (\url{https://github.com/IDSCETHZurich/klask_hardware}), inference software stack at \url{https://github.com/IDSCETHZurich/klask_software} and the simulation and MARL setup at \url{https://github.com/IDSCETHZurich/klask_rl}.
\end{abstract}

\IEEEpeerreviewmaketitle

\section{Introduction}
\label{sec:introduction}

Robot Learning and Multi-agent reinforcement learning (MARL) have achieved remarkable milestones in digital domains. From the superhuman coordination of AlphaGo~\cite{silver2016mastering}, OpenAI Five in Dota 2~\cite{berner2019dota} and AlphaStar in StarCraft II~\cite{vinyals2019grandmaster}, and foundational algorithms like MADDPG~\cite{lowe2017multi}, agents have demonstrated that sophisticated long-horizon behaviors can emerge from simple reward signals given sufficient competitive pressure. 
These digital successes suggest a promising future for embodied intelligence. While these principles are well-established in digital environments, a critical challenge remains: the transition from abstract strategic reasoning to the high-stakes constraints of high-speed physical multi-robot play. A key barrier to this transition is infrastructural. Bridging the reality gap requires a benchmark that balances high-fidelity simulation with accessible hardware. 
While software environments enable millions of interactions per hour, physical MARL is often constrained by a lack of accessible high-fidelity benchmarks. 
Platforms in this domain are frequently either prohibitively expensive industrial systems~\cite{deepmind_tabletennis2024} or limited to single-agent tasks.

\begin{figure}
    \centering
    \includegraphics[width=1\linewidth]{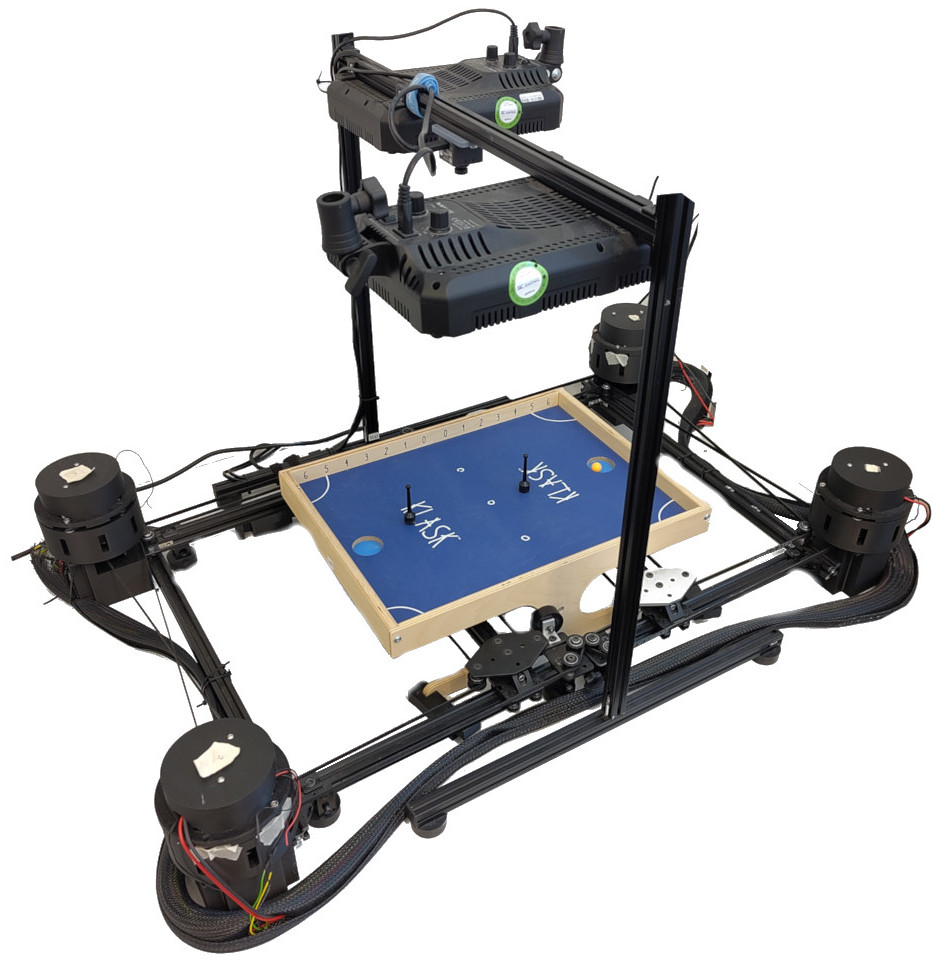}
    \caption{KlaskTron robot with the 2x 2D gantry for each player}
    \label{fig:klasktron}
\end{figure}

We introduce \textbf{KlaskTron} (Fig.~\ref{fig:klasktron}), a platform designed to bridge this gap. KlaskTron is an open-source, low-cost (\textless\$2500 USD), desktop-scale robotic testbed explicitly built for physical adversarial MARL. It occupies a unique niche: more dynamic than static manipulation benchmarks, more accessible than industrial robotics, and designed for two-agent competitive interaction.

\textbf{Why KLASK?} KLASK instantiates a problem class we term \emph{high-speed, contact-rich adversarial MARL}: a zero-sum, continuous-control task in which two agents must simultaneously attack, defend, and cope with explosive contact dynamics under long-horizon credit assignment. Unlike robotic air hockey or table tennis, KLASK requires no compressed-air infrastructure and fits on a standard desk, enabling fully automatic episode resets and continuous physical training without human intervention. Each ${\sim}31$-second episode contains on average 12.4 dynamic contacts (defensive blocks plus offensive strikes).
 With more than one million units sold worldwide, the rules are immediately intuitive to a broad audience, lowering the barrier to demonstration and community engagement.

Our contributions are:
\begin{enumerate}
    \item \textbf{Mechanisms \& Designs:} We provide complete CAD files, bill of materials (BOM), and assembly instructions for a dual-gantry robotic system costing under \$2500 USD with an estimated build time of 20 hours.

    \item \textbf{Modeling and Simulation:} A GPU-native digital twin component for NVIDIA Isaac Lab enabling massively parallel training, capable of processing 20M steps in $\sim$12 minutes on a single RTX 4090.

    \item \textbf{A Carefully Engineered Sim-to-Real Baseline:} A reproducible PPO-based reference agent, comprising shaped-reward design and a learned neural actuator model, that achieves zero-shot transfer to hardware. We release trained weights and the full training pipeline. Out-of-the-box reinforcement learning approaches required substantial reward shaping and replay-buffer engineering to produce competent policies; the released baseline reflects this engineering effort and is offered as a reproducible starting point rather than a definitive algorithmic claim.

    \item \textbf{Full Open-Source Release:} All hardware designs, simulation code, and training pipelines are publicly available to serve as a community-driven research ecosystem.
\end{enumerate}

By providing this complete ecosystem, we establish KlaskTron as a challenging and accessible testbed designed to foster progress at the intersection of MARL, robotics, and the sim-to-real problem.
\section{Related Work}
\label{sec:related_work}

We contribute to the achievements of state-of-the-art robotic games with KlaskTron, an accessible platform that complements current systems and facilitates physical adversarial learning.

\subsection{Collaborative and Fleet Benchmarks}
Recent initiatives have standardized physical benchmarks for multi-robot coordination. 
M4Bench~\cite{yoshida2025m4bench} establishes a framework for Human-Robot Interaction (HRI) manipulation, while FORMIGA~\cite{yalcinkaya2025formiga} addresses fleet management in field robotics. 
Similarly, RoboVerse~\cite{geng2025roboverse} proposes a unified platform for scalable robot learning.
Digital benchmarks like PettingZoo~\cite{tuyls2020pettingzoo} and VMAS~\cite{li2022vmassu} have standardized soft-body and vector simulations.
While these platforms advance cooperative logistics, they do not address the high-frequency, contact-rich adversarial dynamics fundamental to competitive MARL.

\subsection{Single-Agent Dynamic Platforms}
Advances in dynamic physical skill learning have largely focused on single-agent settings. The \textit{CyberRunner} benchmark~\cite{bi2024sample} showcased the potential for low-cost, open-hardware platforms to facilitate physical reinforcement learning for dynamic tasks. While it provides a structured framework for evaluating single-agent performance, it is inherently limited by its design to individual tasks.

At the high-performance end of the spectrum, D'Ambrosio et al.~\cite{deepmind_tabletennis2024, d2023robotic} established the gold standard for high-speed actuation with their robotic table tennis system, demonstrating human-level rallying. However, such systems rely on industrial-grade hardware inaccessible to most labs. 
KlaskTron aims to provide a reference baseline and an accessible platform to this class of problems, leveraging the GPU-parallelized simulation paradigm championed by recent benchmarks like ManiSkill3~\cite{tao2024maniskill3} to enable rapid policy iteration on accessible hardware.

\subsection{Adversarial Tabletop Games}

Robotic game-playing has a rich history, from early foosball systems~\cite{weigel2005kiro} to autonomous pool~\cite{nierhoff2016billiards}, with early adversarial multi-robot competitions such as RoboFlag~\cite{1239093} laying conceptual groundwork for distributed control under competitive constraints. RoboCup~\cite{rebelo2023robocup} remains the premier event for physical multi-agent sports.
The closest functional analog to our work is Robot Air Hockey~\cite{liu2022airhockey}, as surveyed by general game-playing reviews. Recent work by Jankowski et al.~\cite{jankowski2024distilling} has advanced this domain by distilling offline optimal control into reactive policies for high-speed contact planning. KlaskTron complements this control-first literature by providing a learning-first testbed specifically designed for adversarial MARL. This is critical for studying emerging phenomena such as coordinated adversarial attacks~\cite{lee2025wolfpack} and safe multi-agent control~\cite{zhang2025solving}, which require reproducible physical environments to validate theoretical robustness guarantees.

\subsection{Sim-to-Real Methodology}
The reality gap remains a primary bottleneck for physical reinforcement learning~\cite{chebotar2019closing}, as surveyed by Zhao et al.~\cite{zhao2020sim}. To demonstrate the feasibility of the KlaskTron platform, we provide a baseline implementation using established techniques in Domain Randomization and System Identification. Crucially, we adopt the approach of learning a neural actuator model to capture complex motor dynamics (friction, latency) that analytical models like MuJoCo~\cite{todorov2012mujoco} miss. This methodology was recently validated by Fey et al.~\cite{fey2025bridging} demonstrated that unsupervised actuator networks are essential for bridging the sim-to-real gap in athletic loco-manipulation. KlaskTron serves as an accessible testbed where such emerging sim-to-real methodologies can be benchmarked under the unique distribution shifts presented by high-speed adversarial interaction.

\section{The KlaskTron Platform}
\label{sec:platform}

This section describes the KlaskTron hardware, software ecosystem, and the design principles that guided its development.

\subsection{Design Philosophy}

KlaskTron was designed according to three principles for accessibility and reproducibility, following best practices from recent open-source robotics platforms~\cite{bi2024sample,moos2024foosball}:

\begin{enumerate}
    \item \textbf{Accessibility:} Total cost under \$2500 USD using off-the-shelf components and 3D-printed parts.
    \item \textbf{Reproducibility:} Complete open-source release of all designs, with detailed assembly instructions.
    \item \textbf{Research Utility:} Sufficient dynamics complexity to serve as a meaningful benchmark for emergent MARL behaviors. KlaskTron isolates specific open problems in physical RL: high-speed contact dynamics, non-linear reward landscapes, and adversarial intent inference.
\end{enumerate}

\subsection{Hardware Architecture}

\begin{figure*}[!t]
    \begin{center}
        \input{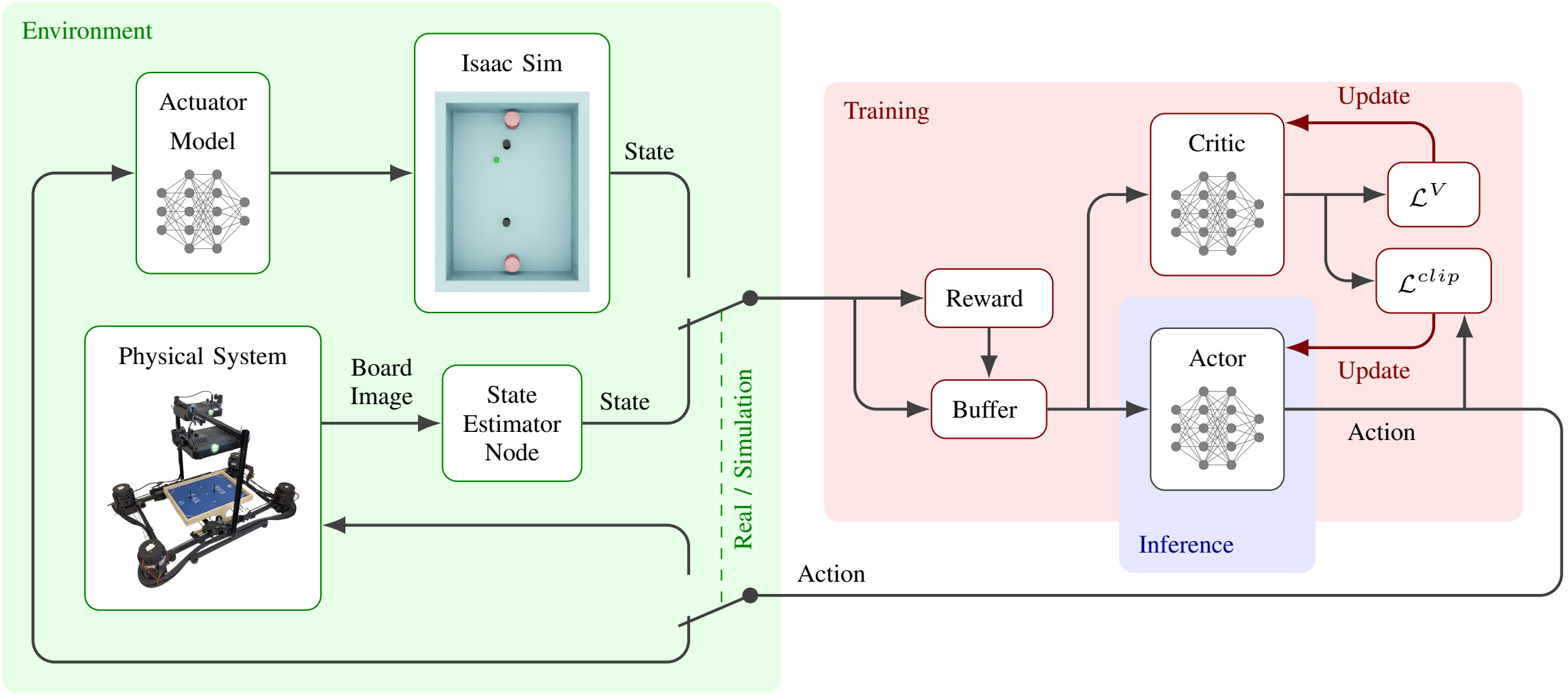}
    \end{center}
    \caption{System architecture of the KlaskTron platform. The environment is interchangeable between the Isaac Lab simulation (for training) and the physical system (for deployment). State estimation in simulation is provided directly by the physics engine; on hardware, an Extended Kalman Filter processes camera observations. The control stack uses PPO during training and the frozen actor network during inference. Actions pass through the learned actuator model in simulation or directly to physical motors on hardware.}
    \label{fig:system-diagram}
\end{figure*}

The KlaskTron hardware (Fig.~\ref{fig:klasktron}) consists of:

\textbf{Dual CoreXY Gantries:} Two identical, opposing CoreXY gantry systems are mounted beneath a standard KLASK game board. The CoreXY architecture is ideal for this application because it keeps the motors stationary, minimizing the inertia of moving components and enabling accelerations (mean $6.25~\text{m/s}^2$) sufficient to traverse the playing field in under a second, matching the pace of human reaction.

\textbf{Direct-Drive Brushless DC Motors:} The system uses direct-drive brushless DC motors controlled by ODrive S1 drives, providing the torque and acceleration necessary for competitive, human-like play. Integrated encoders enable closed-loop velocity control with a measured tracking RMSE of $0.013~\text{m/s}$. We initially evaluated NEMA~17 and sensored stepper motors but observed unavoidable stalls, jerk, and slip that prevented the explosive movements competitive play demands; after comparing several BLDC options we selected the ODrive S1 stack for its reliable sensored control and tuning flexibility, which together unlock the full skill range required by the task.

\textbf{Magnetic Coupling:} Following the original KLASK game design, strikers on the board surface are magnetically coupled to carriages beneath the board, enabling non-contact actuation.

\textbf{Overhead Camera:} A single camera operating at 120 Hz provides observations of the game state. This sensing rate was selected to balance cost-effectiveness with the performance required for high-speed interaction. By providing observations at this frequency (50 Hz), the platform requires agents to account for system latency, which encourages the development of robust predictive representations.

\textbf{Communication:} Low-level motor control is handled via direct CAN bus connection, with a ROS node converting high-level velocity commands to motor controller messages.

\subsection{GPU-Native Digital Twin}

A critical component of the KlaskTron ecosystem is the \textbf{digital twin} implemented in NVIDIA Isaac Lab. This GPU-native simulation environment enables:

\begin{itemize}
    \item \textbf{Massively Parallel Training:} Thousands of independent KLASK environments run simultaneously on a single GPU.
    \item \textbf{Rapid Iteration:} 20 million PPO steps complete in approximately 12 minutes on an RTX 4090.
    \item \textbf{High-Fidelity Physics:} The PhysX engine models board geometry, contact dynamics, and friction with tunable parameters.
\end{itemize}

\begin{figure}[ht]
    \centering
    \includegraphics[width=1\linewidth]{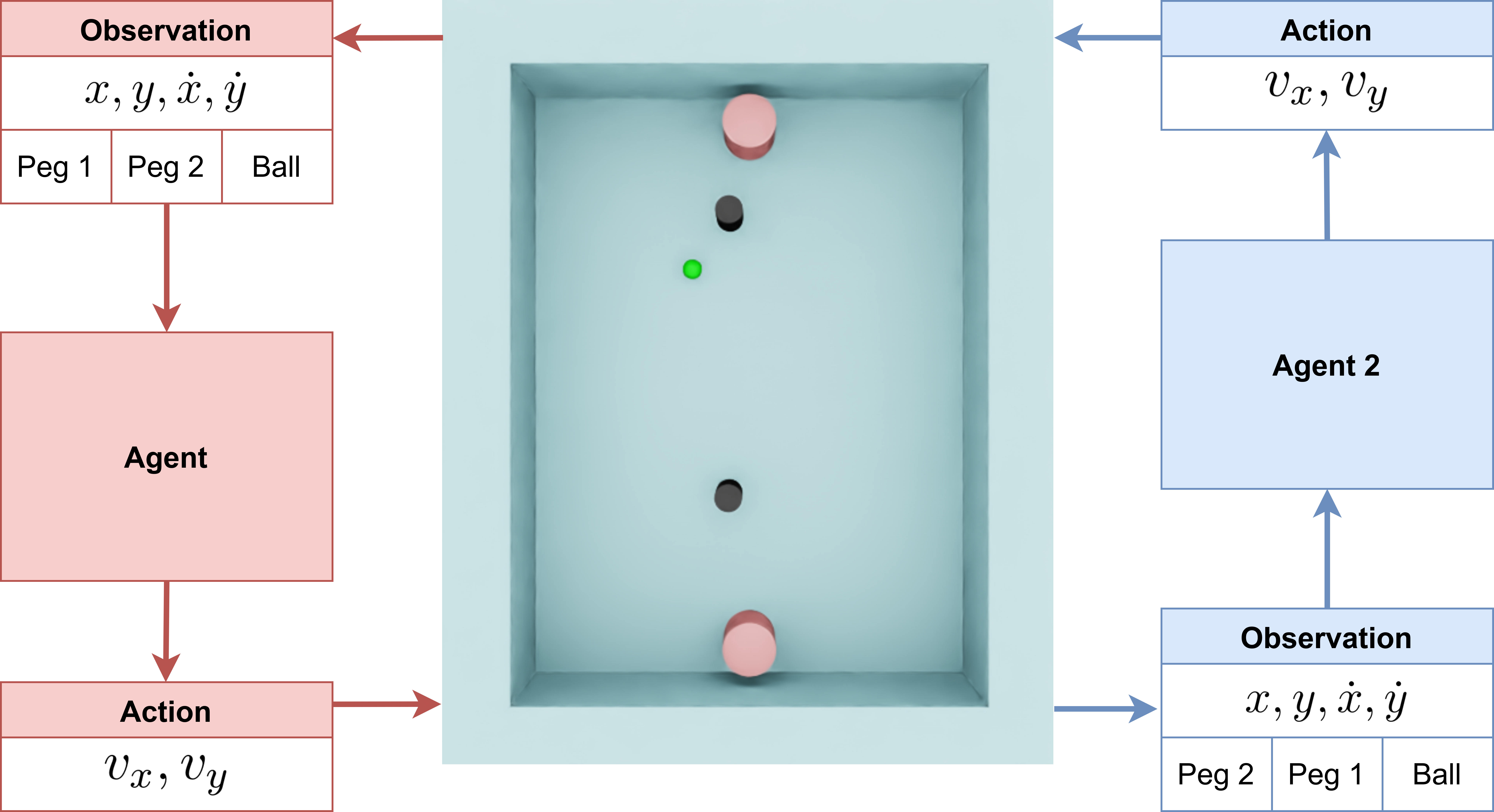}
    \caption{The Isaac Lab simulation environment for KlaskTron. Each agent observes ball and striker positions/velocities and outputs continuous velocity commands.}
    \label{fig:issac_env}
\end{figure}

The simulation (Fig.~\ref{fig:issac_env}) models the board, walls, strikers, and ball using primitive shapes. For the baseline experiments, we simplify the game by omitting the magnetic obstacle pieces (biscuits) present in the commercial game.

\section{Methodology: Baseline Learning and Transfer}
\label{sec:methodology}

To establish the utility of KlaskTron as a research platform, we provide a reference sim-to-real baseline. The primary objective of this methodology is to serve as a reference baseline, verifying that the physical environment's complexity is manageable within current reinforcement learning and system identification paradigms. Rather than defining the platform's limits, these methods provide a validated foundation from which the community can build, iterate, and compare novel algorithmic approaches. 

The following subsections describe the state estimation, actuator modeling, and training protocols used to achieve initial zero-shot policy transfer.

\subsection{Vision-Based State Estimation}

On the physical system, the game state is estimated from camera images using an Extended Kalman Filter (EKF) \cite{julier1997new}. The state vector for each tracked object is defined as $\mathbf{x} = [x, y, \dot{x}, \dot{y}]^T$. To handle the high-speed dynamics and sudden transitions during play, we utilize an \textbf{adaptive process noise} mechanism, a technique established for the identification of variances in filtering systems \cite{mehra1970identification}. 

As illustrated in Fig.~\ref{fig:ekf-matrices}, the process noise covariance $Q_k$ is adjusted dynamically based on the ball's predicted position. In the 
open areas of the board $Q_k$ is small, reflecting high confidence in the physics-based transition model. Near the Boundaries and Strikers regions $Q_k$ increases significantly due to collisions, causing the filter to rely more heavily on real-time measurements to account for non-linear collision dynamics. This enables rapid convergence to post-collision trajectories.

\definecolor{qball_blue}{RGB}{0,0,139}
\definecolor{qball_green}{RGB}{144,238,144}
\definecolor{qball_cyan}{RGB}{0,255,255}
\definecolor{qball_purple}{RGB}{128,0,128}
\definecolor{beigeborder}{RGB}{245,245,220}

\begin{figure}[ht]
\centering
\resizebox{\columnwidth}{!}{%
  \begin{tabular}{@{}c@{\hskip 4mm}c@{}}
    \begin{minipage}[c]{0.55\textwidth}
      \centering
      \begin{tikzpicture}[font=\sffamily, scale=1.0, transform shape]
          \fill[beigeborder] (-2.3, -3.8) rectangle (2.3, 3.8);
          \fill[qball_green] (-2.2, -3.7) rectangle (2.2, 3.7);
          \fill[qball_cyan] (-2.2, -3.6) rectangle (-2.0, 3.6);
          \fill[qball_cyan] (2.0, -3.6) rectangle (2.2, 3.6);
          \fill[qball_purple] (-2.2, -3.7) rectangle (-2.0, -3.6);
          \fill[qball_purple] (2.0, -3.7) rectangle (2.2, -3.6);
          \fill[qball_purple] (-2.2, 3.6) rectangle (-2.0, 3.7);
          \fill[qball_purple] (2.0, 3.6) rectangle (2.2, 3.7);
          \fill[qball_blue] (-2, -3.6) rectangle (2, 3.6);
         
          \fill[white] (0, 3.2) circle (0.3);
          \fill[white] (0, -3.2) circle (0.3);
          \fill[qball_purple] (0, 1.5) circle (0.4);
          \fill[black] (0, 1.5) circle (0.2);
          \fill[qball_purple] (0, -1.5) circle (0.4);
          \fill[black] (0, -1.5) circle (0.2);
          \node[white, scale=1.5] at (0, 2.3) {KLASK};
          \node[white, scale=1.5, rotate=180] at (0, -2.3) {KLASK};
          \draw[white, fill=white] (-1, 0) circle (0.15);
          \draw[white, fill=white] (0, 0) circle (0.15);
          \draw[white, fill=white] (1, 0) circle (0.15);
          \fill[yellow] (0.5, -0.7) circle (0.1);
         
          \coordinate (axis_origin) at (2.3, 3.8);
          \draw[->, thick, black] (axis_origin) -- ++(-0.7, 0) node[above] {$y$};
          \draw[->, thick, black] (axis_origin) -- ++(0, -0.7) node[right] {$x$};
      \end{tikzpicture}
    \end{minipage}
    &
    \begin{minipage}[c]{0.35\textwidth}
      \textcolor{qball_blue}{Blue region:}
      \[
      Q_{\text{ball}} = \operatorname{diag}(2.0, 2.0, 30.0, 30.0)
      \]
     
      \textcolor{qball_green}{Green region:}
      \[
      Q_{\text{ball}} = \operatorname{diag}(6.0, 2.0, 1030.0, 30.0)
      \]
     
      \textcolor{qball_cyan}{Cyan region:}
      \[
      Q_{\text{ball}} = \operatorname{diag}(2.0, 6.0, 30.0, 1030.0)
      \]
     
      \textcolor{qball_purple}{Purple region:}
      \[
      Q_{\text{ball}} = \operatorname{diag}(6.0, 6.0, 1030.0, 1030.0)
      \]
    \end{minipage}
  \end{tabular}%
}
\caption{Adaptive process noise regions on the KLASK board (left) with corresponding covariance matrices (right). Higher values in $Q$ matrices indicate reduced confidence in the process model, causing the EKF to rely more on measurements.}
\label{fig:ekf-matrices}
\end{figure}
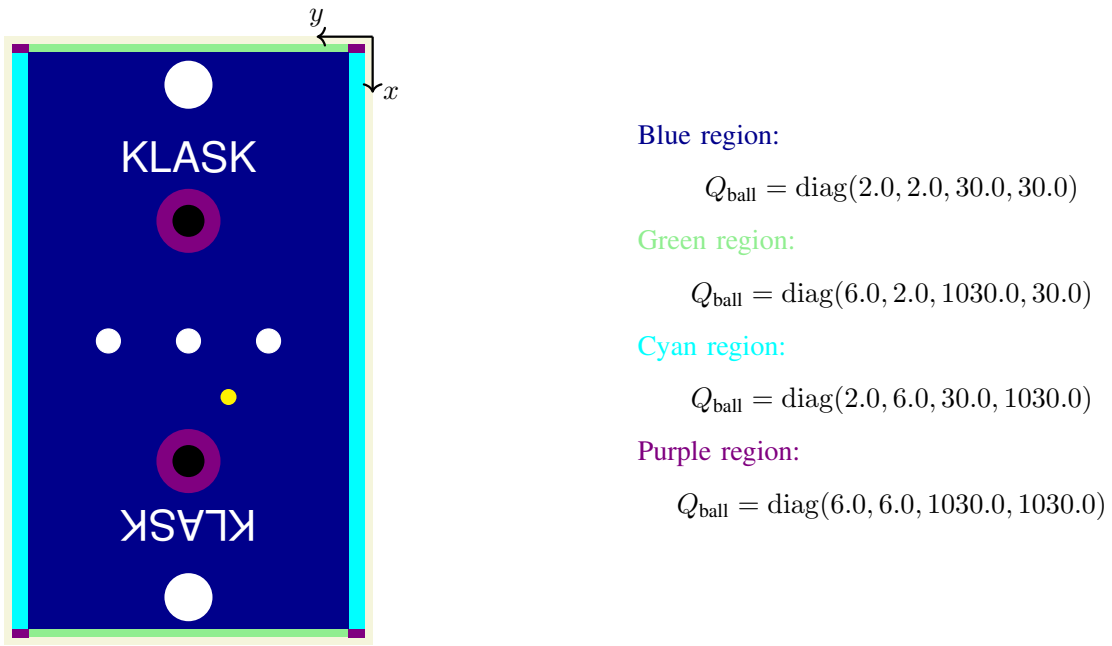

\subsection{The Sim-to-Real Challenge}

Training directly on hardware is impractical due to the sample inefficiency of reinforcement learning, necessitating simulation-based training. The platform's high-speed dynamics expose the reality gap in ways that static or slower manipulation tasks do not: idealized Proportional--Derivative controllers, even when augmented with standard domain randomization, produce control signals that the hardware cannot faithfully execute. The system must account for non-linearities such as friction, the torque--speed envelope of the brushless motors, and the magnetic coupling between the under-board carriage and the surface peg, together with the transient delays inherent in a 50\,Hz control loop and the adversarial distribution shifts induced by an unpredictable opponent. By providing a testbed where idealized simulation fails, KlaskTron offers a rigorous environment for validating sophisticated transfer mechanisms such as the learned neural actuator model we describe next.

\subsection{Learned Neural Actuator Model}
\label{sec:actuator_model}

Following prior work on learned actuator dynamics~\cite{actuatorpolicy2019}, we address the sim-to-real gap by replacing the idealized PD actuator in simulation with a learned model that captures real motor behavior. \textbf{Data Collection:} We collected 37,500 time-series samples ($\sim$12.5 minutes at 50 Hz) from the physical system, executing diverse trajectories (circles, diagonals, linear sweeps) with commanded velocities up to $\pm 0.7$ m/s. \textbf{Model Architecture:} A three-hidden-layer MLP (64 units each) predicts the next velocity from a history of commands and velocities:
$$v_{t+1} = f_{NN}(v_t, a_{t-1}, u_{t-1}, \dots)$$
The architecture and history length were selected by a trajectory-based ablation across $\{1, 5, 10\}$-step history and $\{32, 64, 128\}$-unit widths trained over 8 random seeds; the 10-step / $3{\times}64$ configuration achieved the lowest velocity-prediction RMSE while remaining compact, and is the model used for all results reported in this paper.

\textbf{Deconstructing the Sim-to-Real Gap:} The 10-timestep input history enables the model to implicitly learn system latency and transient response, performing action interpolation without explicit time-delay buffers. By training on trajectories with velocities up to $\pm 0.7$ m/s, the network learned the physical torque--speed envelope, identifying saturation points where commanded accelerations are physically unachievable. We established this model as the primary transfer mechanism by demonstrating successful zero-shot deployment with Domain Randomization disabled, whereas policies relying on Domain Randomization with idealized physics failed to transfer.

\begin{figure}[h]
    \centering
    \includegraphics[width=1\linewidth]{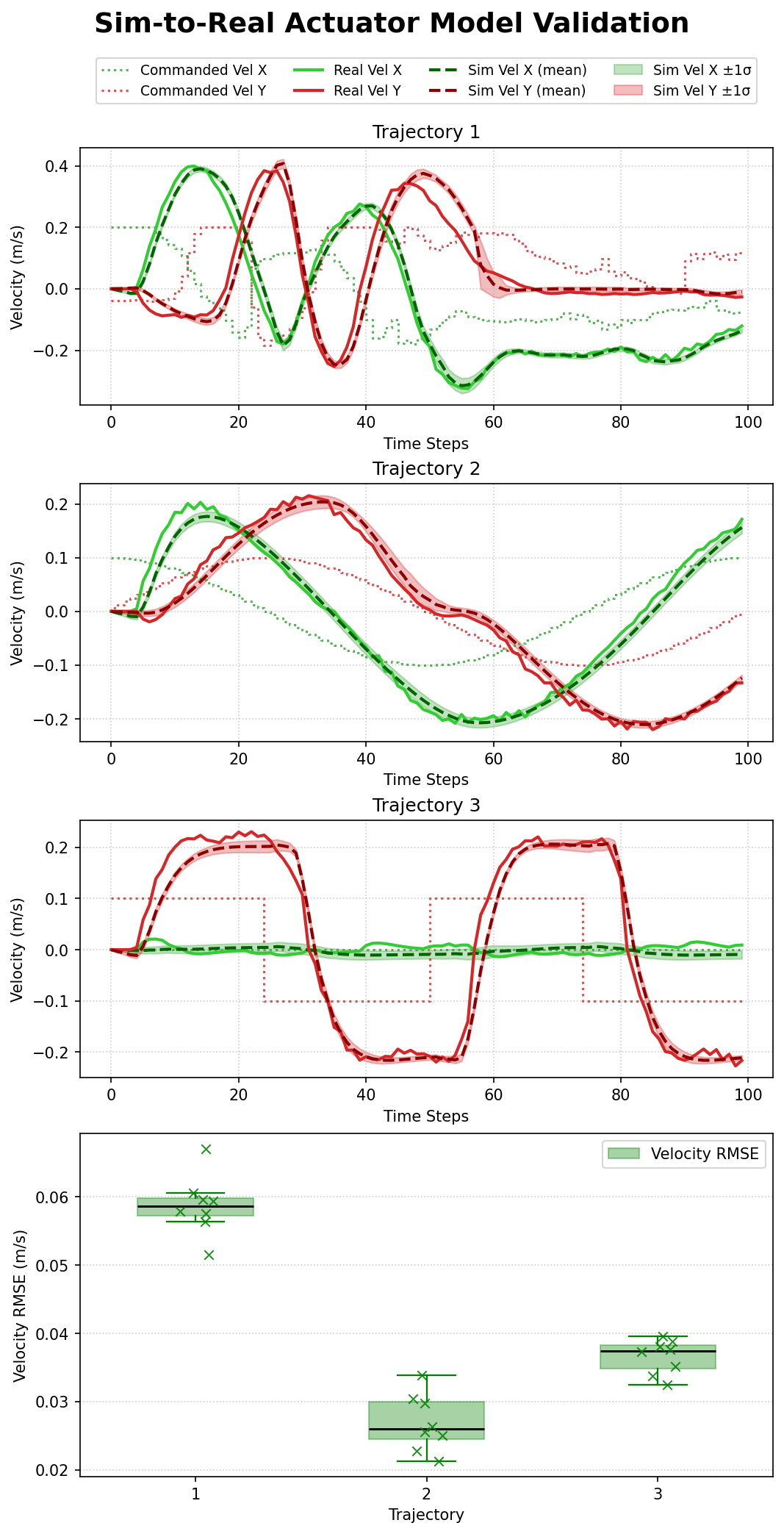}
    \caption{Validation of the learned actuator model (10-step history, three hidden layers of 64 units). Solid lines show real hardware response; dashed lines show simulated response to identical command sequences. Shaded bands show the spread across 8 training seeds, indicating that prediction quality is robust to initialisation. The model achieves a global validation RMSE of $0.013~\text{m/s}$, with peak errors of $0.032~\text{m/s}$ (x) and $0.049~\text{m/s}$ (y) on these representative high-frequency trajectories.}
    \label{fig:actuator_validation}
\end{figure}

Figure~\ref{fig:actuator_validation} validates the model: simulated velocities closely track real system response across diverse trajectories, and the seed-band collapses tightly around the mean prediction.

\subsection{Training Protocol}

Out-of-the-box reinforcement learning approaches required substantial reward shaping and replay-buffer engineering to produce competent policies on this task; the PPO baseline described below reflects this engineering effort and is offered as a reproducible starting point rather than a definitive algorithmic claim. We summarise the components here; full reward formulae, curriculum schedules, and self-play hyperparameters are provided in Appendix~\ref{app:training_protocol}.

\textbf{Reward and Curriculum.} Sparse game-outcome rewards alone yielded passive policies. We therefore use a time-decaying curriculum that initially guides exploration with dense shaping signals (offensive progress, defensive positioning, ball analysis) and transitions to sparse adversarial incentives (goals scored, conceded, and self-goal penalties).

\textbf{Self-Play.} Adversarial competence is developed via Enhanced Self-Play~\cite{berner2019dota}: the agent trains against a pool of past checkpoints sampled with geometric decay favouring recent policies.

\subsubsection{MDP Formulation}
The KLASK game is modeled as a two-player zero-sum Markov Decision Process (MDP) defined by the tuple $(\mathcal{S}, \mathcal{A}, \mathcal{R}, \mathcal{P}, \gamma)$.

\textbf{State Space ($\mathcal{S}$):} The observation vector consists of two components:
\begin{itemize}
    \item \textit{Standard Observations:} Positions and velocities of the player's striker, the opponent's striker, and the ball.
    \item \textit{Augmented Features:} Relative angle between the striker and the goal, Euclidean distance from the ball to the goals, and distance from the striker to the ball.
\end{itemize}

\textbf{Action Space ($\mathcal{A}$):} Each agent outputs continuous planar velocity commands:
$$a_t = (v_x, v_y) \in [-0.4, 0.4]^2~\text{m/s}$$

\textbf{Objective:} Each agent maximizes expected discounted rewards:
\begin{equation}
\pi^* = \arg\max_{\pi} \mathbb{E}_{\pi} \left[ \sum_{t=0}^{\infty} \gamma^t r_t \right]
\end{equation}
where $\gamma$ is the discount factor (see Appendix~\ref{app:hyperparams}).

\subsection{Evaluation Metrics}

We report agent quality through three complementary metrics standardised in our benchmark protocol: \emph{scoring rate} (goals per minute of physical play), \emph{episode length} (seconds per goal), and \emph{contact frequency} (defensive blocks plus offensive strikes per episode). For agent-versus-agent comparisons in simulation we additionally use the Elo rating system~\cite{berner2019dota} with a $K$-factor of 10 and baseline rating of 1200. Detailed Elo derivations and round-robin curricula results are provided in Appendix~\ref{app:training_protocol}.

\section{Experiments: Validating the Testbed}
\label{sec:experiments}

We present experiments demonstrating that KlaskTron functions as a reproducible physical benchmark: simulated policies transfer zero-shot to hardware, the platform is reliable across extended evaluation, and the residual gap between agent and human play is well characterised. Supporting algorithmic analyses (curriculum comparison, self-play variants, observation augmentation) are reported in Appendix~\ref{app:training_protocol}.

\subsection{Sim-to-Real Transfer}
\label{sec:sim2real}
The utility of the KlaskTron benchmark depends on the feasibility of transferring simulated policies to physical hardware. Our experiments identify a clear dependency between actuator modelling and transfer success.

\textbf{Actuator Modelling Constraints.} Initial attempts using idealised Proportional--Derivative controllers with domain randomization failed to transfer: the discrepancy between simulated and physical command response produced control signals the hardware could not execute, leading to immediate policy failure.

\textbf{System Identification and Transfer.} Incorporating the learned actuator model (Section~\ref{sec:actuator_model}, Fig.~\ref{fig:actuator_validation}) into the training pipeline enables successful zero-shot transfer. While domain randomization is a common sim-to-real heuristic, our high-fidelity actuator model allowed transfer even with DR disabled, suggesting that for KlaskTron precise system identification is a more critical factor than broad parameter randomization.

\textbf{Emergent Behaviours.} Transferred policies exhibited differentiated offense/defense (autonomous switching between defensive positioning and offensive strikes), goal-directed play (consistent scoring against stationary and random opponents), and reactive adaptation to ball trajectories and opponent movements.

\subsection{Hardware Failure Modes}
\label{sec:failure_modes}
The residual $0.5$--$1.5~\text{cm}$ offensive misses observed in human-robot play are policy and perception artefacts rather than hardware failures. The dominant physical effect is the magnetic coupling between the under-board carriage and the surface peg, which lags by approximately $0.4~\text{cm}$ during direction changes; combined with gantry damping and EKF position uncertainty this produces a non-trivial sub-centimetre offset that the policy must learn to anticipate. Quantitatively, decelerating the peg from its peak velocity of $0.5~\text{m/s}$ to standstill consumes 77\,ms and 2\,cm, while accelerating it back to the same speed in the opposite direction takes 111\,ms and 7\,cm. Kinematic constraints intrinsic to the magnetically coupled actuator that even human players experience and that the agent must internalise rather than overcome.

\subsection{System Reliability and Repeatability}
\label{sec:reliability}
\textbf{Hardware reliability.} Across evaluations totalling more than 30 consecutive games we observed no hardware-level failure, where hardware failure is defined as motor connection error, camera dropout, EKF divergence, peg decoupling, or ball departure from the board. Software-level resets occurred and were logged but did not corrupt the protocol.

\textbf{Software reproducibility.} Two students independently retrained the baseline policy on different days using the released training pipeline; both reached comparable competitive levels in physical play, providing initial evidence that the released hardware--software stack constitutes a repeatable substrate for RL evaluation.

\textbf{Sources of stochasticity.} We deliberately distinguish \emph{system reliability} (faithful command execution; velocity-tracking RMSE $0.013~\text{m/s}$) from \emph{system stochasticity} (magnetic coupling slip and minor mechanical variation) and \emph{policy stochasticity} (action sampling noise). Natural system stochasticity is preserved by design, since the research question is whether learned policies cope with real-world randomness, not eliminate it. Our released benchmark protocol formalises scoring rate, episode length, and contact frequency as the primary reporting metrics.

\subsection{Human-Robot Benchmark Validation}
\label{sec:human_robot}
To evaluate the platform's potential as a high-ceiling benchmark we tested the baseline policy against a recreational human opponent over a 10-game series (Table~\ref{tab:human_robot}). The agent achieved a 10\% win rate; each ${\sim}31$-second episode contained on average 12.4 dynamic contacts (defensive blocks plus offensive strikes), confirming that physical adversarial play emerges in the policy. The agent exhibited several emergent strategies, including reactive goal-tending and the use of board geometry for defensive bank shots.

Performance was limited by the offensive action mismatch characterised in Section~\ref{sec:failure_modes}: the policy often selected the correct velocity vectors but the striker missed the ball by $0.5$--$1.5~\text{cm}$. Together with the $31.3~\text{s}$ average episode duration, this confirms substantial research headroom remains on the KlaskTron platform.

\begin{table}[h]
\centering
\caption{10-Game Human-Robot Tournament Statistics. The agent demonstrates emerging defensive competence but significant offensive limitations.}
\label{tab:human_robot}
\resizebox{\columnwidth}{!}{%
\begin{tabular}{@{}llccccc@{}}
\toprule
\textbf{Game} & \textbf{Winner} & \textbf{Duration (s)} & \textbf{Defensive} & \textbf{Offensive} & \textbf{Goal-Directed} & \textbf{Possession} \\
\textbf{\#} & & & \textbf{Blocks} & \textbf{Strikes} & \textbf{Hits} & \textbf{(\%)} \\ \midrule
1 & Human & 28 & 6 & 4 & 1 & 42\% \\
2 & Human & 35 & 8 & 5 & 2 & 45\% \\
3 & \textbf{Robot} & 42 & 12 & 9 & 4 & 51\% \\
4 & Human & 21 & 4 & 3 & 1 & 38\% \\
5 & Human & 30 & 7 & 5 & 2 & 44\% \\
6 & Human & 28 & 6 & 4 & 1 & 43\% \\
7 & Human & 25 & 5 & 3 & 1 & 40\% \\
8 & Human & 32 & 9 & 6 & 2 & 46\% \\
9 & Human & 29 & 6 & 4 & 1 & 43\% \\
10 & Human & 43 & 11 & 7 & 3 & 44\% \\ \midrule
\textbf{Avg.} & \textit{Human (90\%)} & \textit{31.3} & \textit{7.4} & \textit{5.0} & \textit{1.8} & \textit{43.6\%} \\ \bottomrule
\end{tabular}%
}
\end{table}

\section{Conclusion}
\label{sec:conclusion}

We have introduced KlaskTron, an open-source platform that lowers the infrastructural barrier to physical adversarial multi-agent reinforcement learning. The contribution is the platform itself: hardware designs, a GPU-native digital twin, a formalised gameplay protocol with reproducible reliability statistics, and a carefully engineered PPO-based reference baseline released as trained weights and a complete training pipeline. The accompanying experiments characterise the platform rather than advance an algorithmic claim. They quantify hardware reliability over $30^+$ consecutive games, attribute the residual sub-centimetre offensive miss to physical effects the policy must learn (Section~\ref{sec:failure_modes}), and isolate the actuator-modelling components most critical for zero-shot transfer.

Future work will extend the platform along two axes: increased task complexity through reintroduction of the magnetic biscuit obstacles, yielding a non-convex landscape that demands multi-dimensional strategy, and natural extension to four-agent play via the commercially available round-Klask variant, opening team-based MARL questions on the same hardware. Beyond these, we aim to move past hand-engineered features by exploring autonomous representation learning and to integrate game-theoretic approaches that approximate Nash equilibria in competitive physical environments. Uniquely positioned between static manipulation tasks and industrial-grade systems, KlaskTron is intended as a community substrate on which future algorithmic work in physical MARL can be developed and compared.

\section*{Acknowledgments}
The authors thank the IDSC technical staff Dani Wagner and Matthias Mueller for their constant support and help in developing the hardware, and Thomas Bi for his support and consultation on various topics regarding reinforcement learning and hardware development.

\bibliographystyle{IEEEtranN}
\bibliography{references}

\onecolumn
\newpage
\appendix

\subsection{Training Protocol: Reward, Curriculum, Self-Play, Observations}
\label{app:training_protocol}

\subsubsection{Reward Curriculum}
The total reward at time $t$ combines a sparse termination signal with time-decaying shaping rewards:
\begin{equation}
r_{total}(t) = r_{term} + \sum_i R_i \cdot \exp\left(-\frac{t}{N}\right) \cdot r_{shape,i}
\end{equation}
where $N$ is the decay time constant.

\textbf{Termination ($r_{term}$):} Goal Scored ($+1$), Goal Conceded ($-1$), Agent Self-Goal Error ($-1$).

\textbf{Shaping ($r_{shape}$):}
\begin{itemize}
    \item \textit{Offensive Progress:} $\exp(-\|p_B - p_{O}\|/D)$ (ball to opponent goal)
    \item \textit{Defensive Positioning:} $-\exp(-\|p_B - p_{P}\|/D)$ (ball to own goal)
    \item \textit{Ball Analysis:} $\exp(-\|p_{PS} - p_B\|/C)$ (proximity) and $\|v_B\|$ (speed)
\end{itemize}

The decay schedule encourages meaningful interaction during initial training while progressively removing the training wheels to prevent reward hacking. Sparse-only reward variants yielded passive policies in our preliminary experiments; the curriculum was selected from a comparison of three variants (collision-only, dense ball-to-goal, sparse-only), with the dense ball-to-goal curriculum reaching the highest Elo in round-robin tournaments (Elo $\approx 1283$ vs. $\approx 1240$ for sparse and $\approx 1220$ for collision-only).

\subsubsection{Self-Play Configuration}
Adversarial competence is developed via Enhanced Self-Play~\cite{berner2019dota}: the agent trains against a pool of past checkpoints sampled with geometric decay, $w_i = 0.9^{(N - i - 1)}$, where $i$ indexes checkpoints and $N-1$ is the most recent. The opponent pool updates when the average score difference exceeds $0.7$. We compared this against random pool sampling and best-opponent-only sampling: the best-opponent variant proved unstable due to insufficient opponent diversity, while Enhanced Self-Play and random pool sampling produced comparable competent policies.

\subsubsection{Observation Augmentation}
Following findings in large-scale game domains~\cite{berner2019dota}, we provide hand-engineered features (striker-to-ball vector, angle to goal) alongside the raw state vector. Augmented observations accelerate long-term convergence relative to raw-state baselines.

\subsubsection{PPO Hyperparameters}
\label{app:hyperparams}

\begin{table}[h]
\centering
\begin{tabular}{ll}
\toprule
\textbf{Parameter} & \textbf{Value} \\
\midrule
Discount factor ($\gamma$) & 0.99 \\
GAE parameter ($\lambda$) & 0.95 \\
Learning rate & $3 \times 10^{-4}$ \\
Clip parameter ($\epsilon$) & 0.2 \\
Mini-batch size & 4096 \\
Entropy coefficient & 0.01 \\
\bottomrule
\end{tabular}
\caption{PPO hyperparameters.}
\end{table}

\subsection{Domain Randomization}

For the best-performing transferred policy, domain randomization was \textbf{disabled}. The following ranges were used in ablation studies:

\begin{table}[h]
\centering
\begin{tabular}{ll}
\toprule
\textbf{Parameter} & \textbf{Range} \\
\midrule
Static friction & $[0.2, 0.5]$ \\
Dynamic friction & $[0.3, 0.45]$ \\
Restitution & $[0.95, 0.99]$ \\
Ball mass & $[1.8, 2.2]$ mg \\
\bottomrule
\end{tabular}
\caption{Domain randomization ranges (disabled for best agent).}
\end{table}

\subsection{Actuator Model Details}
\label{app:actuator}

\subsubsection{Data Collection Protocol}

Training data was collected by executing scripted trajectories:
\begin{itemize}
    \item Circular paths (varying radii and speeds)
    \item Diagonal sweeps
    \item Linear velocity ramps
    \item Random walk sequences
\end{itemize}

Total: 69,669 samples at 50 Hz ($\sim$23.2 minutes). Commanded velocities ranged $\pm 0.7$ m/s.

\subsubsection{Model Architecture}

\begin{itemize}
    \item Input: History of 10 timesteps (commands and velocities)
    \item Hidden layers: 3 $\times$ 64 units, ReLU activation
    \item Output: Predicted velocity $\hat{v}_{t+1}$
    \item Training: MSE loss, Adam optimizer, 1000 epochs
\end{itemize}

Validation RMSE on held-out trajectories: 0.013 m/s (global average). The architecture and history length were selected by an ablation over $\{1, 5, 10\}$-step history $\times$ $\{32, 64, 128\}$-unit widths trained over 8 random seeds; the 10-step / $3{\times}64$ configuration achieved the lowest combined velocity-prediction RMSE while remaining compact.

\end{document}